\documentclass[sigconf]{acmart}

\AtBeginDocument{%
  \providecommand\BibTeX{{%
    \normalfont B\kern-0.5em{\scshape i\kern-0.25em b}\kern-0.8em\TeX}}}

\setcopyright{none}

\usepackage{xspace}
\usepackage{csquotes}

\newcommand\ie{i.\,e.\xspace}
\newcommand\eg{e.\,g.\xspace}

\newcommand\abs[1]{| #1 |}

\newcommand{\fairalg}{\textsf{ALD}\xspace} 
\usepackage{colortbl}
\usepackage{enumitem}
\usepackage[ruled,vlined,linesnumbered]{algorithm2e}
\SetKwInput{cstInput}{input}
\newlength\myindent
\usepackage{mathtools}
\DeclarePairedDelimiter{\ceil}{\lceil}{\rceil}
\usepackage{subcaption}
\usepackage{booktabs}
\usepackage{pbox}
\usepackage{}
\usepackage{dcolumn}
\newcolumntype{L}{D{.}{.}{2}}

\begin{document}

\title{Locating disparities in machine learning}

\author{Moritz von Zahn}
\email{vzahn@wiwi.uni-frankfurt.de}
\affiliation{%
  \institution{Goethe University Frankfurt}
  \streetaddress{Theodor-W.-Adorno-Platz 4}
%  \city{Frankfurt am Main}
  \country{Germany}
  \postcode{60323}
}

\author{Oliver Hinz}
\email{hinz@wiwi.uni-frankfurt.de}
\affiliation{%
  \institution{Goethe University Frankfurt}
  \streetaddress{Theodor-W.-Adorno-Platz 4}
%  \city{Frankfurt am Main}
  \country{Germany}
  \postcode{60323}
}

\author{Stefan Feuerriegel}
\email{feuerriegel@lmu.de}
\affiliation{%
  \institution{LMU Munich}
  \streetaddress{Geschwister-Scholl-Platz 1}
%  \city{Munich}
  \country{Germany}
}

\renewcommand{\shortauthors}{von Zahn, Hinz, Feuerriegel}

\begin{abstract}
Machine learning can provide predictions with disparate outcomes, in which subgroups of the population (\eg, defined by age, gender, or other sensitive attributes) are systematically disadvantaged. In order to comply with upcoming legislation, practitioners need to locate such disparate outcomes. However, previous literature typically detects disparities through statistical procedures for when the sensitive attribute is specified a priori. This limits applicability in real-world settings where datasets are high dimensional and, on top of that, sensitive attributes may be unknown. As a remedy, we propose a data-driven framework called \emph{Automatic Location of Disparities} (\fairalg) which aims at locating disparities in machine learning. \fairalg meets several demands from industry: \fairalg (1)~is applicable to arbitrary machine learning classifiers; (2)~operates on different definitions of disparities (\eg, statistical parity or equalized odds); (3)~deals with both categorical and continuous predictors even if disparities arise from complex and multi-way interactions known as intersectionality (\eg, age above 60 \emph{and} female). \fairalg produces interpretable audit reports as output. We demonstrate the effectiveness of \fairalg based on both synthetic and real-world datasets. As a result, we empower practitioners to effectively locate and mitigate disparities in machine learning algorithms, conduct algorithmic audits, and protect individuals from discrimination.
\end{abstract}




\maketitle
\pagestyle{plain}

\section{Introduction}

Machine learning (ML) is nowadays widely used by companies and organizations. However, ML is known to be subject to bias (c.f. \cite{Barocas.2019}). This refers to disparities in which outcomes of ML systematically deviate from statistical, moral, or regulatory standards \citep{Danks.2017}, especially in ways that disadvantage people from certain sociodemographics (gender, race, or other attributes deemed sensitive). For example, ML in criminal justice has been found to return output with systematic disadvantages towards black defendants \citep{Angwin.2016}. Specifically, the {COMPAS} algorithm falsely labels black defendants as \textquote{high risk} more frequently than white defendants, potentially leading to a disproportionate number of black defendants being kept in prison. Systematic disadvantages have been found in various industry applications of ML, such as lending \citep{Hardt.2016}, marketing \citep{vonZahn.2021}, and automated hiring \citep{Raghavan.2020}.

Companies are increasingly under pressure to assess potential disparities in their ML applications due to legal concerns. For example, in June 2023, the European Parliament approved the AI~Act that specifically demands \textquote{bias monitoring [and] detection} \cite{EU.2023} for multiple types of ML applications in industry. Comparable legal initiatives are emerging worldwide, including the National New Generation Artificial Intelligence Governance Expert Committee in China \cite{Roberts.2021} and the Algorithmic Accountability Act in the US \cite{US.2022}. In many cases, these initiatives extend from and reinforce existing anti-discrimination laws. For example, the Equal Credit Opportunity Act in the US demands decisions to grant a loan to be independent of attributes such as age, gender, race, or religion \citep{Congress.1974}. Similarly, the Equal Employment Opportunity Commission in the US aims at preventing discrimination in hiring decisions, thereby limiting the extent to which disparate outcomes may arise when ML is used for this purpose \citep{Barocas.2016}.

In industry applications of ML, detecting disparate outcomes is non-trivial because the sensitive attributes inducing such disparities are not given a~priori. On top of that, disparities can even arise from complex, multi-way interactions where only small subsets of individuals are affected (\eg, age above 60 \emph{and} female). With the rise of big data, the number of potentially sensitive attributes can be high and the possible combinations grow exponentially. In addition, for sensitive attributes that are continuous (\eg, age), there is the question of how to find suitable splits (\eg, are disparities present for age above 30, 40, 50, \ldots years?). These issues are particularly relevant in light of the recent debate on intersectionality (cf. \cite{Collins.2020}). Intersectionality refers to cases where a combination of different predictors is linked disparate outcomes (but where such disparities are not present when considering them separately). Importantly, several cases of intersectionality have documented in recent years. For example, ``hidden'' disparities were driving hiring decisions at General Motors \citep{Crenshaw.2015}. In this example, discrimination by either race or gender alone was not present. Both black workers and women were hired at large rates; \ie, black workers (men) primarily for the production line and women (white) as clerks. However, almost no black women were hired, demonstrating discrimination against this subgroup that eventually resulted in a lawsuit. Another example concerns healthcare, where ML has labeled female Hispanic patients as \textquote{healthy} despite illness at a significantly higher rate than other patients \citep{Seyyed.2021}. As a consequence, female Hispanics have potentially delayed access to care, putting their physical well-being at risk. 

Disparate outcomes are commonly measured via fairness metrics. Various fairness metrics have been designed to \emph{quantify} the magnitude of disparities when the subgroups are defined a~priori (\eg, the absolute odds difference to quantify disparities with regard to equalized odds \citep{Hardt.2016}). Given a fairness metric and a subgroup, prior research has put forth algorithms to detect disparate outcomes (\eg, \cite{Black.2020}). However, these algorithms cannot \emph{locate} sensitive attributes (and thus the corresponding disadvantaged subgroup). For continuous variables (\eg, age), it means that a certain criterion (\eg, age above 55) has to be identified, for which the subgroup is subject to disparate outcomes. A simple brute-force search is computationally intractable for high-dimensional datasets, as well as for continuous variables (see Sec.~\ref{sec:related_work} for details). 

\textbf{Proposed framework:}\footnote{Our \fairalg framework is open source and available as an R package. URL: \url{https://github.com/moritzvz/ald}.} In this work, we propose a novel, data-driven framework called \emph{Automatic Location of Disparities} (\fairalg). \fairalg aims at locating subgroups of the population that are subject to disparate outcomes. For this, it overcomes the above-mentioned limitations of existing algorithms: it does not need a~priori knowledge of which attribute is involved, but locates the affected sensitive attributes (within the existing predictors) in a data-driven manner. It can even find subgroups due to intersectionality where complex, multi-way interactions among predictors introduce disparities. To achieve this, we combine conditional inference trees \citep{Hothorn.2006, Strobl.2009} with statistical hypothesis testing. In the first step, we randomly split the data into two halves. Using the first half of the data, we employ randomized conditional inference trees to recursively partition the space of sensitive attributes to find statistically significant associations with the disparity measure of interest. Using the second half of the data, we then evaluate the leaves (\ie, subgroups) of the trees via $\chi^2$ hypothesis testing, corrected for multiple hypothesis testing, and then rank the results. As output, \fairalg produces interpretable reports that can be used for algorithmic audits. We demonstrate the effectiveness of \fairalg using both synthetic and real-world datasets.

\textbf{Practical value:} A main advantage of \fairalg is its broad applicability. Specifically, \fairalg can be used to audit arbitrary machine learning classifiers, consider different notions of fairness, and handle both categorical and continuous predictors in low- or high-dimensional settings. Moreover, \fairalg performs subgroup generation and evaluation based on separate data and accounts for multiple hypotheses testing, thus providing valid confidence measures. Importantly, it also identifies disparities due to intersectionality where disparities arise from complex, multi-way interactions only, as has been shown to occur in practice (see above). Altogether, \fairalg helps industry practitioners to locate disparities in ML algorithms so that disparate outcomes can be mitigated. \fairalg thus supports practitioners in adhering to upcoming legislation and protecting individuals from discrimination.

\noindent
\textbf{Contributions:} The main contributions of our work are:
\begin{enumerate}[leftmargin=0.5cm]
    \item We propose a novel, data-driven framework for automatically locating disparate outcomes (\fairalg) in machine learning. \fairalg is open source\textsuperscript{1} and carefully crafted based on the needs of companies and organizations. 
    \item \fairalg provides interpretable fairness reports and accurately locates disparities even in complex cases (\eg, due to \textquote{hidden} intersectionality).
\end{enumerate}

\section{Related work}
\label{sec:related_work}

\subsection{Notions of fairness in ML}

Disparities are closely related to the concept of fairness in ML, which refers to systematic deviations in ML outcomes \citep{Danks.2017, Kleinberg.2018, Corbett.2018, Barocas.2019}. Over recent years, several fairness notions have been developed (cf. \cite{Barocas.2019, DeArteaga.2022} for an overview). Here, fairness notions are oftentimes described at, \eg, the level of individuals or the level of groups. 

The former, (a)~individual fairness, stipulates that individuals with similar properties also receive similar outcomes \citep{Dwork.2012, Ranzato.2021}. Individual fairness thus relies upon a concept of similarity to make comparisons when individuals are disadvantaged compared to others. However, in practice, it is often unclear how to define similarity; because of this, it is difficult for practitioners to come up with population-wide interpretations where fairness is violated.

The latter, (b)~group-level fairness, stipulates that no disparities should occur for the outcomes of individuals inside and outside of a group $G$ \citep{Dwork.2012, Hardt.2016}. This group is typically identified by a sensitive attribute, such as, for instance, gender, age, and race \citep{Barocas.2016, Kleinberg.2018}. If group-level fairness is violated, one group will then receive outcomes to their disadvantage. Without loss of generality, such a group can also be a subgroup that is described by a combination of multiple attributes, which is referred to as intersectionality \citep{Collins.2020}. A benefit of group-level fairness is that the disadvantaged subgroup can be easily recognized through a sensitive attribute, thereby facilitating interpretability. Motivated by this, our work also focuses on group-level fairness to identify disparities, and we thus review different metrics from group-level fairness (\eg, statistical parity \citep{Dwork.2012}, equalized odds \citep{Hardt.2016}) in the following.

\subsection{Metrics for measuring disparities in ML}

Within the area of group-level fairness, there is no `universal' definition of fairness; instead, several metrics $\psi$ may be used to measure disparate outcomes \citep{Corbett.2018}. The exact choice depends on the legal framework \citep{Barocas.2019} and the context of the application such as the business goals \citep{vonZahn.2021, Feuerriegel.2022}. Specifically, the assessment of to what extent disparities are acceptable (\ie, $\psi(G, \mathcal{X} \setminus G) \leq \psi^\ast$ for a threshold $\psi^\ast$) is a case-by-case decision that should be carefully made by industry practitioners. Under certain mathematical conditions, it is even impossible for several of them to be simultaneously satisfied \citep{Kleinberg.2017, Pleiss.2017, Friedler.2016, Chouldechova.2017compas}. 

Example metrics for measuring disparities are as follows. \textbf{Statistical parity} requires that the outcome $Y$ is independent of the sensitive attribute(s) $a$ \citep{Dwork.2012}, \ie, that the likelihood of outcome $y \in Y$ is equal across a subgroup $G$ (\eg, young women) and outside of it ($\mathcal{X} \setminus G$). Formally, one can measure such disparities via the statistical parity difference
\begin{equation} \label{eq:sp}
    \psi_\text{sp} = P(Y = y \mid G) - P(Y = y \mid \mathcal{X} \setminus G) . 
\end{equation}
Statistical parity is highly relevant, as it plays a crucial role in many legal frameworks \cite{Barocas.2019}. However, a main drawback is that it does not account for cases in which correlations between sensitive attributes and outcomes are deemed acceptable. In such cases, a viable alternative may be equalized odds. \textbf{Equalized odds} requires that the error rates (\ie, false positive rate $\mathit{FPR}$ and false negative rate $\mathit{FNR}$) are independent of $a$ \citep{Hardt.2016}. Put differently, the likelihood that the ground truth is predicted correctly shall be the same across a subgroup and outside of it. Formally, such disparities are quantified via the absolute odds difference
\begin{equation} \label{eq:eo}
    \psi_\text{eo} = \frac{1}{2}\,\big[\lvert\mathit{FPR}_\mathrm{G}-\mathit{FPR}_\mathrm{\mathcal{X} \setminus G}\rvert + \lvert\mathit{FNR}_\mathrm{G} - \mathit{FNR}_\mathrm{\mathcal{X} \setminus G}\rvert\big] .
\end{equation}
For a detailed survey on fairness metrics, we refer to \citep{Barocas.2019}. 

Disparities may also be assessed across combinations of sensitive attributes to prevent so-called \textquote{fairness gerrymandering} \citep{Kearns.2018, Asudeh.2020}. For example, \citep{Kearns.2018} propose definitions of disparities that aim to reconcile group-level and other individual notions of fairness. The authors show that both assessing and efficiently mitigating these disparities can be modeled as finding the Nash equilibrium in a zero-sum game between ML classifier and auditor. Importantly, \citep{Kearns.2018} illustrate the profound computational challenges that arise when assessing disparities across subgroups of sensitive attributes and rely on linear regression for doing so. Notably, linear modeling techniques have limitations, potentially leading to non-linear patterns going unnoticed.

A related stream in the literature seeks to identify `unwarranted' associations in ML classifiers (\eg, \cite{Tramer.2017, Black.2020}). For instance, FlipTest by \citep{Black.2020} detects if individuals had been treated differently given an alternate group membership. For this, an optimal transport map takes the distribution of a given sensitive attribute to generate a distribution of counterparts, so that the outcomes of both can be compared. However, FlipTest can only detect disparate outcomes manually, \ie, solely for sensitive attributes that are specified a~priori. This implies that the practitioner must manually determine which groups to consider in the analysis, which is highly subjective and prone to manipulation (\eg, by intentionally leaving out groups). The same limitation of having a~priori knowledge of the affected sensitive attributes also underlies FairTest by \citep{Tramer.2017}, fairness-based permutation by \citep{DiCiccio.2020}, and tests based on counterfactual fairness by \citep{Kusner.2017}. As a consequence, the aforementioned tools can merely confirm (or reject) and quantify disparities that were previously defined by a practitioner. Importantly, they do not provide industry practitioners with an effective, broadly applicable tool to \emph{locate} the sensitive attributes (\ie, the subgroups affected). The latter is our contribution.

\subsection{Tools for locating disparities in ML}

In what follows, we review tools for auditing ML so that disparities are located. Here, the objective is, given a trained ML classifier, to locate attributes that induce disparate outcomes.

One approach for locating disparities would be to perform an exhaustive search across all potential combinations of sensitive attributes (and their values) and, for each, test for the presence of disparities. However, such exhaustive search would be impractical. Let us assume $k$ sensitive attributes $A = \{a_1, a_2, \ldots, a_k\}$, each is categorical with $W_{a_1}, \ldots, W_{a_k}$ different levels. Then, one would need to test the powerset overall all levels of all sensitive attributes; that is, there are a total of $\prod_{a \in A} (2^{W_a}-1)$ tests. For instance, in the context of credit lending under the Equal Credit Opportunity Act \citep{Congress.1974}, one would need to inspect 9 different sensitive attributes. Assuming that each attribute has five descretized values, the total number would be $\approx 38 \cdot 10^9$. Hence, for high-dimensional datasets, exhaustive search is prohibited.

In order to avoid an exhaustive search, \citet{Zhang.2016} have developed a subset scan method to reveal predictive bias, \ie, subsets of the population for which an underlying classifier exhibits over- or underestimation. In certain cases, this method is able to locate disparity-inducing attributes in high-dimensional and complex settings. However, the method has two important limitations. First, it only locates sensitive attributes affected by over- or underestimation, and thus is not suited for other forms of disparities (such as statistical parity or equalized odds). Second, it searches the space of sensitive attributes greedily and thus fails to locate disparities that are due to intersectionality (\ie, remain hidden when considering the sensitive attributes separately). Both limitations are addressed by our data-driven framework \fairalg.

\citet{Chouldechova.2017} introduce a model-comparison framework for comparing \emph{two} classifiers and, for this, build upon trees that allow for splitting sensitive attributes. Specifically, the tree compares two classifiers in terms of an arbitrary measure of group-level fairness. It greedily splits the space of sensitive attributes based on statistical testing and aims to maximize the disparity in the child nodes. Thereby, an exhaustive search of all possible subsets is avoided, but disparities due to intersectionality may remain undetected. In addition, the framework is designed for comparing the output of \emph{two} classifiers, and, hence, the procedure cannot directly be used for auditing a dataset or ML application.

\textbf{Research gap:} As we shall see later in our experiments, the accuracy of the above tools is limited, especially for high-dimensional data and continuous attributes. In particular, none of the abovementioned tools is able to systematically locate disparities, especially not in the presence of intersectionality. These shortcomings are addressed in our work, and, to this end, we propose a novel, data-driven framework: \emph{Automatic Location of Disparities} (\fairalg).

\section{Framework for automatic location of disparities}

\subsection{Task description}

We study the setting where an ML algorithm is audited for disparate outcomes. Our task is to locate disparities based on logged data of a given ML classifier. Specifically, we want to identify subgroups of the population who are advantaged (or disadvantaged) at an overproportionate rate with regard to some fairness notion and thus lead to disparities in ML outcomes. In our task, we want to produce output that is interpretable. That means, we want to infer a criterion characterizing individuals in the population where disparities occur. An example of such criterion could be ``age $\geq 60$ \emph{and} gender = female''. Owing to this, we refer to our task to as \emph{locating} disparities.

\textbf{Input:} Our framework builds upon the following input. (1)~Let $D = (x_i, y_i)_i$, $i = 1, \ldots, M$ with $(x_i,y_i) \in \mathcal{X} \times \{ 0, 1 \}$ be some data that should be used during the audit (\eg, logged data from an ML classifier). For terminology, we refer to the different predictors in $\mathcal{X}$ as $a = \{a_k \,\mid\, k = 1, \ldots, K\}$ (\eg, age, gender). We distinguish categorical predictors $a_k^\text{(cat)}$ and continuous predictors $a_k^\text{(cont)}$. (2)~Let $\psi$ be a function for measuring fairness at group level (\eg, statistical parity, equalized odds). It thus quantifies the disparity $\psi(G, \mathcal{X} \setminus G) \in \mathbb{R}$ across a group $G$ vs. outside of it $\mathcal{X} \setminus G$. 

\textbf{Objective:} We aim at finding disjoint subgroups $G_j \subseteq \mathcal{X}$ subject to disparities. That is, there should be different outcomes wrt. the fairness notion $\psi$ within the subgroup ($G_j$) vs. outside of it ($\mathcal{X} \setminus G_j$). We test this empirically by using the available data $D$. For interpretability, the different subgroups are defined by some criterion $c_j$ based on the available predictors $a_k$, that is, $G_j = \{ x \in \mathcal{X} \,\mid\, x \text{ satisfies } c_j \}$, where each criterion is a logical expression of the form
\begin{equation}
\begin{split}
\mathbf{1} \Big( a_{k_1}^\text{(cat)} = \overline{a}_1^\text{(cat)} \Big) \wedge \mathbf{1} \Big( a_{k_2}^\text{(cat)} = \overline{a}_2^\text{(cat)} \Big) \wedge \ldots \\
\wedge \mathbf{1} \Big( a_{k_1}^\text{(cont)} \gtrless \overline{a}_1^\text{(cont)} \Big) \wedge \mathbf{1} \Big( a_{k_2}^\text{(cont)} \gtrless \overline{a}_2^\text{(cont)} \Big) \wedge \ldots ,
\end{split}
\label{eq:criterion}
\end{equation}
where $\mathbf{1}$ is the indicator function. If desired, one can use only a subset of the available predictors in the logical expression, so that \fairalg performs the analysis with regard to only a few ``sensitive'' attributes. 

\textbf{Challenges:} The problem of finding subgroups associated with disparate outcomes in an ML classifier is daunting for a number of reasons. (1)~Due to intersectionality, the number of possible subgroups grows exponentially in the number of predictor (as well as their cardinality if the predictor is categorical). (2)~Handling predictors that are continuous is non-trivial. The reason is that continuous predictor need to be mapped onto a simple yet interpretable criterion (\eg, a binary criterion for interpretability). (3)~Disparities might be subject to Simpson's paradox \citep{Simpson.1951}. Specifically, disparities may occur in specific subgroups, even when there are no disparities at the overall population level, or the disparities may even appear in the opposite direction. Such patterns have been frequently observed in real-world decision-making. For instance, graduate admissions typically do not reveal a gender bias when the gender bias is assessed at a university level, yet a strong gender bias becomes evident at the department level (\eg, where there are more men in engineering and more women in arts) \citep{Bickel.1975}. (4)~Data-driven subgroup generation and subsequent evaluation may yield misleading confidence measures (\eg, in the form of downward distorted $p$-values). For example, if subgroups are generated based on the same data used to test the underlying hypothesis later, the false discovery rate will be inflated \citep{Andrews.2019, Kuchibhotla.2022}.

\subsection{Overview of \fairalg framework}

In order to locate disparity-inducing attributes, we develop our \fairalg framework in the following. \fairalg proceeds along three steps: 
\begin{itemize}[leftmargin=0.5cm]
\item[(1)] \textbf{Recursive partitioning for subgroup generation.} \fairalg splits the data into halves $D_1$ and $D_2$, whereby $D_1$ serves as the basis for binary recursive partitioning. The recursive partitioning is initialized multiple times, with each tree considering a random subset of $D_1$. The recursive partitioning can be viewed as a search tree, where, within a search tree, each node splits a random subset of attributes so that the disparity $\psi$ between the resulting subgroups is maximized. The terminal nodes of the trees are returned, as these provide the candidate subgroups $\mathcal{G}$. Each of the candidate subgroups $G_j \in \mathcal{G}$ is then defined by a criterion $c_j$ (\ie, a logical expression) that is obtained by concatenating the partitioning operations from each node in the search tree via logical ``and''s.

\item[(2)] \textbf{Statistical hypothesis testing for subgroup-specific disparity assessment.} For each subgroup, \fairalg then performs statistical testing based on $D_2$ to check for the presence of subgroup-specific disparities. Formally, this step tests each subgroup $G_j \in \mathcal{G}$ if it is subject to disparate outcomes with regard to some fairness metric $\psi$. For this, the ML outcomes from all logged data $D$ in the subgroup $G_j$ is compared to the ML outcomes from all data $D$ outside of the subgroup ($\mathcal{X} \setminus G_j$). We then use $\chi^2$-tests in order to perform statistical hypotheses testing. The test statistics examines whether the subgroup $G_j$ receive disparate outcomes with large statistical confidence. We account for the problem of multiple hypothesis testing by applying Benjamini-Hochberg correction \citep[see, \eg,][]{Bogdan.2008}.

\item[(3)] \textbf{Audit report generation.} \fairalg ranks all subgroups with statistically significant disparities. This is only done for the purpose of ease-of-use so that practitioners receive a prioritization based on which they can examine disparities in manual, downstream analyses. For instance, we expect practitioners to seek potential root causes of the disparities in the decision logic of the ML classifier, in the training data, or, in the data collection practice. Hence, the prioritization should help efficiently allocating resources for manual investigations.

In \fairalg, practitioners may choose between two ranking mechanisms: (i)~\emph{Confidence ranking}. The ranking can be done by considering the statistical confidence in that there are disparities (\ie, the $\chi^2$-test statistic of the hypothesis test). Here, \fairalg ranks disparities higher where it has a larger statistical confidence that these are non-zero. (ii)~\emph{Magnitude ranking}. The ranking can also be based on the magnitude of disparity $\psi$. 

Subsequently, the ranking of choice is combined into an audit report (supported by appropriate visualizations). For each subgroup, the audit shows, \eg: the criterion defining the subgroup (\eg, ``age above 60 and female''), the number of individuals in the group, the magnitude of disparity, and the $\chi^2$-test statistic.

\end{itemize}
The following sections detail each of the three steps in \fairalg.

\subsection{Recursive partitioning for subgroup generation (Step 1)}

In Step~(1), \fairalg begins by randomly splitting the data $D$ into separate halves $D_1$ and $D_2$. $D_1$ is used in this step and serves as the basis for subgroup generation, whereas $D_2$ is kept for the subsequent step to then evaluate the magnitude and confidence of the associated disparities. This separation ensures valid estimates in the final audit report. \fairalg proceeds by performing a binary recursive partitioning of $D_1$ and, thereby, producing subgroups that become increasingly more granular. For this, it builds upon a random forest \citep{Strobl.2009} of conditional inference trees (\ie, ctree \citep{Hothorn.2006}). Let $N_\text{trees}$ denote the number of search trees in the forest, $rnd(D)$ a random sample of data $D$ without replacement, and $rnd(a)$ a random subset of attributes $a$ without replacement. Each search tree recursively splits an independently drawn $rnd(D)$ into smaller subgroups. At each node (\ie, subgroup) in a search tree, the following steps are performed: (i)~The  subset of attributes $rnd(a)$ is randomly drawn. (ii)~Permutation tests are performed to select the split attribute $a_k \in rnd(a)$ and, subsequently, choose the binary split on $a_k$ yielding the strongest discrepancy between outcome $y$ of the two resulting subgroups. For this, the permutation tests follow the framework developed by \citet{Strasser.1999} and \citet{Hothorn.2015}. For each $a_k \in rnd(a)$, a separate permutation test based on the hypothesis of independence between $a_k$ and $y$ yields a $p$-value. The statistical test then indicates the association between $a_k$ and $y$ and thus determines the split. (iii)~The recursive split is performed, which yields two child nodes (with two smaller, disjoint subgroups). For the recursive partitioning, the stopping criterion is met if all $p$-values of the associations between attributes $a \in rnd(A)$ and outcome $y$ exceed a predefined threshold~$\alpha$. In this case, no statistically significant association can be found, and the node is terminal. Finally, once the recursive partitioning is completed, the path to each terminal node (\ie, subgroup $G_j$) is retrieved to yield the criterion $c_j$.

The reasoning behind the above recursive split is as follows. Recall that the recursive partitioning in step (1) should automatically locate disparate outcomes, even in complex cases of continuous sensitive attributes or \textquote{hidden} intersectionality. For this, the recursive split is formalized in a way that (a)~exploration and (b)~exploitation are balanced:
\begin{enumerate}[label=(\alph*),leftmargin=0.5cm]

\item On the one hand, complex combinations of sensitive attributes need to be explored even if they do not exhibit disparities when considered separately. This is a consequence of the Simpson's paradox \citep{Simpson.1951} and related phenomena, which lend to cases of intersectionality \citep{Crenshaw.2015, Collins.2020}. Here, a pure greedy search on $D$ would fail to locate such disparities, as each attribute separately is not associated with the outcome. For \fairalg, the need for exploration is met by fitting several randomized trees which differ both in their access to the input data ($rnd(D)$) and their access to the sensitive attributes for each split ($rnd(a)$). As a result, the trees and their terminal nodes are \textquote{forced to be different} and explore the search space more broadly \citep{Strobl.2009}. We later demonstrate that this enables \fairalg to locate disparities due to intersectionality (in contrast to single trees or \textquote{greedy} subset scans).

\item On the other hand, an exponentially growing search space demands an efficient procedure. Therefore, retrieved evidence of disparate outcomes need to be exploited at an early stage. The need for exploitation is addressed within the randomized trees, where a split is chosen greedily, \ie, the attribute most strongly associated with the outcome is selected. As the splitting criterion relies on statistical significance of at least $\alpha$, both the problem of overfitting and the bias in variable selection commonly found for traditional trees (\eg, for CART \citep{Loh.2014}) is avoided. As we shall see later, this helps \fairalg to accurately locate disparities in continuous attributes.
\end{enumerate}

\subsection{Statistical hypothesis testing for subgroup-specific disparity assessment (Step 2)}
\label{sec:analysis_of_di}

In Step~(2), each subgroup $G_j \in \mathcal{G}$ is subject to statistical testing whether it is associated with disparities in the ML outcome $y$. Recall that the statistical testing is based on $D_2$, \ie, independent of the data used in the previous step. Here, we draw upon the corresponding metric $\psi$ to measure disparities across two groups. Reassuringly, $\psi$ can be an arbitrary group-level fairness notion (such as, \eg, statistical parity or equalized odds). Based on $\psi$, we compare the individuals from the current subgroup ($G_j$) against all individuals outside of the group ($\mathcal{X} \setminus G_j$).

Formally, we define a statistical hypothesis 
\begin{equation}
\mathrm{H}_0 : \psi(G_{j},\, \mathcal{X} \setminus G_j) = 0 ,
\end{equation}
which we test via a $\chi^2$-test. The computation of $\chi^2$ is detailed for statistical parity and equalized odds separately in Appendix~\ref{app:chi}. Notably, $\chi^2$ depends on the magnitude of $\psi$ and on the number of samples in $G_j$ (cf. \citep{Greenwood.1996} for an overview on the $\chi^2$-test). Hence, when assessing disparate outcomes, hypothesis testing takes into account that there may be a small number of observations in a subgroup. Importantly, we apply Benjamini-Hochberg correction to account for the problem of multiple hypothesis testing \citep{Glickman.2014}.

\subsection{Audit report generation (Step 3)}

Step~(3)~generates an audit report that is intended to facilitate ML practitioners. For this, \fairalg ranks the subgroups $G$ so that practitioners can prioritize further inspections and thus allocate their resources strategically. Here, \fairalg offers two ranking mechanisms: 
\begin{enumerate}[label=(\alph*), leftmargin=0.5cm]
\item \emph{Confidence ranking.} The subgroups are ranked according to the $p$-value from the $\chi^2$-test (reflecting the departure of the estimated value of a parameter from its hypothesized value relative to its standard error). Formally, this implies that
\begin{equation}
    %\chi^2_{\text{rank}_1} \geq \chi^2_{\text{rank}_2} \geq \ldots \geq \chi^2_{\text{rank}_J} ,
    p_{\text{rank}_1} \leq p_{\text{rank}_2} \leq \ldots \leq p_{\text{rank}_J} 
\end{equation} 
where $J$ is the number of different subgroups and where $p_{\text{rank}_j}$ denotes the $p$-value for the subgroup of rank $j$. Thereby, subgroups are prioritized for which there is large statistical confidence that the disparity is non-zero. Hence, this ranking returns subgroups at the top of the list that have the largest risk of discrimination. This is especially useful in ML practice where companies and organizations face legal risks due to the pure presence of disparities (and not necessarily the underlying magnitude of the disparities).
\item \emph{Magnitude ranking}. An alternative ranking is by the impact of the disparities on the ML outcomes. For this, subgroups are ranked according to the magnitude of the disparity, so that large disparities appear first in the list. Formally, this implies that
\begin{equation}
    \psi_{\text{rank}_1} \geq \psi_{\text{rank}_2} \geq \ldots \geq \psi_{\text{rank}_J} ,
\end{equation}
where $\psi_{\text{rank}_j}$ denotes the magnitude of disparity with regard to $\psi$ for the subgroup with rank $j$. This ranking mechanism may support ML practitioners in locating individuals that have the largest disadvantage due to disparities.
\end{enumerate}
Based on the above ranking, an audit report is compiled listing the subgroups, as well as accompanying information: (i)~the criterion as an interpretable logical expression identifying the subgroup (\eg, ``age above 60 and female''), (ii)~the absolute number and relative share of individuals in the group relative to the population, (iii)~the magnitude of disparity (\eg, statistical parity difference), (iv)~the $\chi^2$-test statistic, and (v)~the corresponding corrected $p$-value (\ie, the probability of observing the given disparity despite independence between ML outcome and group membership).  

\begin{figure}[h]
  \centering
  \includegraphics[scale=0.15]{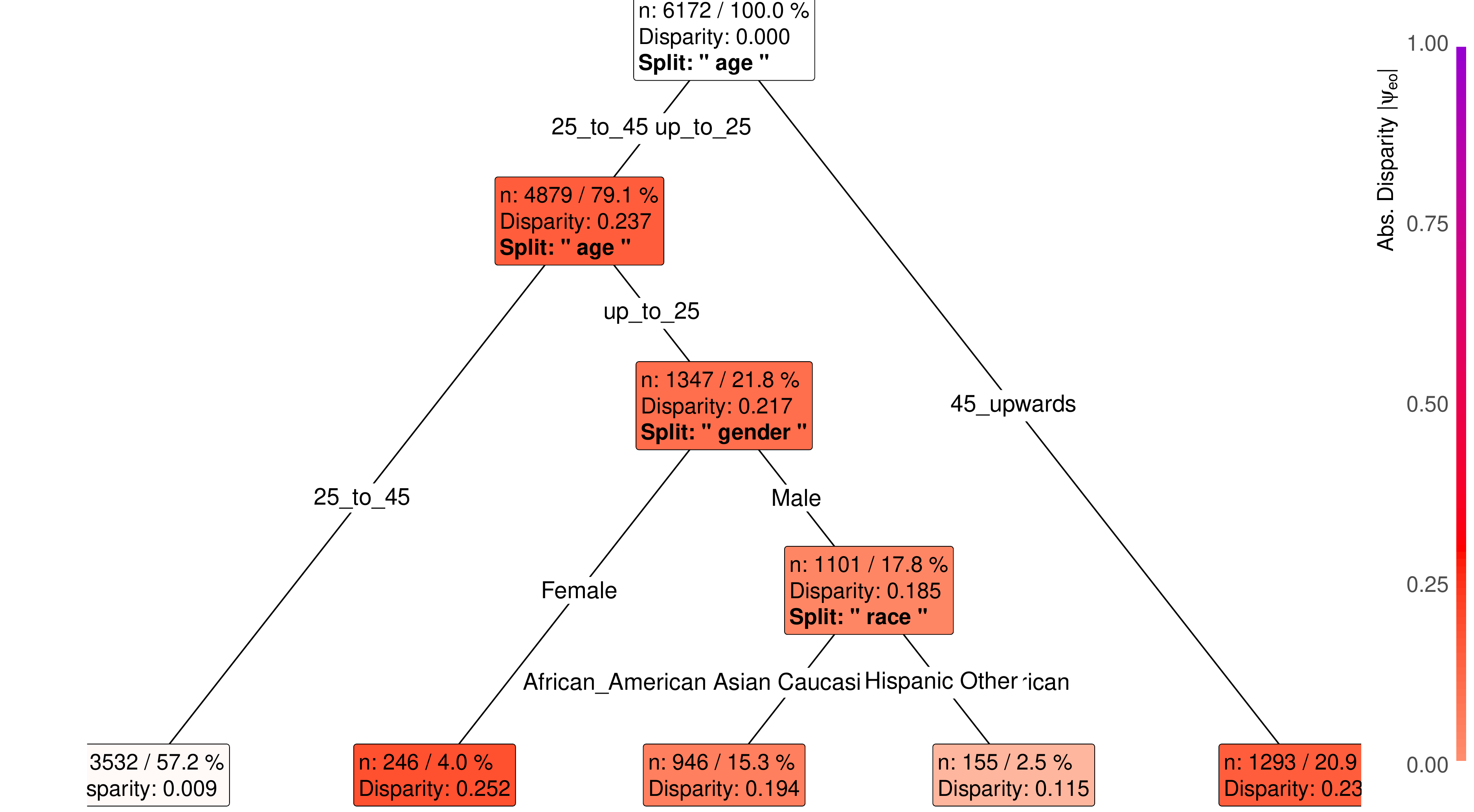}
  \begin{tabular}{p{\columnwidth}}
  \tiny
  Legend: example audit report on the COMPAS dataset \citep{Angwin.2016}. The nodes represent the subgroups of different granularity. Each node shows the number of samples included in the subgroup in absolute and relative terms as well as the magnitude of disparity $\psi$. In addition, the color of the node is determined by the absolute magnitude of disparity $|\psi|$. Within non-terminal nodes, the split variable is given in bold text (\eg, ``race''). For each split, the labels on the edges to the child nodes indicate the respective split values (\eg, ``Hispanic'').
  \end{tabular}
  \caption{Example of a visualization provided by the audit report in \fairalg. }
  \Description{Visualization of an example audit report in \fairalg} % for, e.g.,  blind people
  \label{fig:ggparty_viz}
  \vspace{-0.5cm}
\end{figure}

\textbf{Visualization}: The audit report is accompanied by a visualization of selected search trees that identify subgroups at risk of disparities. The search trees from Step (1) are selected so that the $N$ highest ranking subgroups are shown. In each search tree, the nodes (\ie, the subgroups) are plotted and, moreover, colored according to the magnitude of disparity $\psi$. As a consequence, practitioners can quickly assess both how strong subgroups are affected and how the disparities break down into more granular subgroups. An example plot is shown in Figure~\ref{fig:ggparty_viz}.

\subsection{Implementation} 

% code 
We implemented our \fairalg framework in \texttt{R} and released it as open source R package.\footnote{Our \fairalg framework can be remotely installed from \url{https://github.com/moritzvz/ald}} We implemented the visualization using the packages \texttt{ggparty} and \texttt{ggplot2}. In our implementation, all predictors $a$ that are deemed sensitive by common legal frameworks are scanned for disparities (=\,default). If desired, practitioners can change the default and run \fairalg on a subset $a' \subset a$ that is manually selected to include domain knowledge. 

We carefully configured the parameters for the random forest consisting of $N_\text{trees}$ conditional inference trees \citep{Hothorn.2006, Hothorn.2015, Strobl.2009} in step~(1). Our choices were influenced by both empirical results and established defaults. Specifically, we set $N_\text{trees} = 25$, which we found to perform well in our experiments. We maintained the fraction of the sample $rnd(D)$ at $0.632$, consistent with the default value in \cite{Hothorn.2015}. Similarly, we opted for the default number of attributes in $rnd(a)$, $\ceil{\sqrt{K}}$, a commonly employed value in random forest-based applications. For the subsequent audit report in step~3, we recommend a default value of $N_\text{groups} = 3$ to strike a balance between interpretability and granularity.

\section{Experimental setup}

We evaluate the effectiveness of \fairalg against existing baselines (Sec.~\ref{sec:baselines}). For this, we draw upon two synthetic datasets (Sec.~\ref{sec:synth_dataset}) and two real-world datasets (Sec.~\ref{sec:real_dataset}).\footnote{In all our experiments, \fairalg completes all calculations on a standard desktop computer within seconds.} The purpose of the synthetic datasets is to compare the identified disparities against ground truth and thus assess the overall detection accuracy of the different methods. The purpose of the real-world datasets is to demonstrate the applicability of \fairalg to actual datasets. Here, we compare the identified disparities against domain knowledge. In all experiments, \fairalg is applied directly to the logged data (\ie, to circumvent the need to train an additional, intermediate ML classifier).

\subsection{Baselines}
\label{sec:baselines}

We compare the performance of \fairalg with two other tools for locating disparities\footnote{Notably, we have also compared the performance to that of \textquote{fairness gerrymandering} \cite{Kearns.2018} which we implemented as an adapted version of the code available at \url{https://github.com/algowatchpenn/GerryFair}. However, this method relies on linear models as its foundation, while the synthetic datasets inherently exhibit non-linear disparities. Consequently, none of the disparities were accurately detected using this approach. Thus, we have opted to refrain from presenting these results in Fig.~\ref{fig:res1}.}:
\begin{itemize}[leftmargin=0.5cm]

\item \textbf{Multivariate subset scan} \citep{Zhang.2016}: We adapt the subset scan \citep{Zhang.2016} so that it can be applied in our experiments. The original version is limited to the fairness notion of predictive bias; we thus adapt the scoring function to work with the notion of statistical parity.

\item \textbf{Parameter instability tree} \citep{Chouldechova.2017}: We adapt the parameter instability tree from \citet{Chouldechova.2017} for our experiments. The original version compares \emph{two} classifiers by performing recursive splits using tailored permutation tests based on the conditional inference framework \citep{Hothorn.2015}. In our adapted version, we build upon the same framework but, as we focus on a single dataset, use standard permutation tests instead \citep{Hothorn.2015}. 

\end{itemize}
Of note, other baselines (e.g., FairTest, FlipTest) are not applicable as they cannot \emph{automatically} locate disparate outcomes and require the affected attributes to be specified a~priori.

We later compare the performance of the above baselines against \fairalg ($N_\text{groups} = 3$ with confidence-based ranking). The performance is measured by how often attributes responsible for disparate outcomes are located across 100 random runs. We consider output as successful if (i)~split values in continuous predictors (e.g., young vs. old) are accurately located with $\pm5\,\%$ tolerance, and (ii)~subgroups identified do not feature any condition that includes a non-relevant predictor or sensitive attribute. We evaluate the performance on two difficult synthetic datasets, namely locating disparate outcomes due to a continuous predictor and due to ``hidden'' intersectionality. 

\subsection{Synthetic datasets}
\label{sec:synth_dataset}

\textbf{Synthetic dataset 1:} We sample $n = 10,000$ observations with three predictors (race, gender, and age) and a binary outcome $y$. The samples are generated so that age and/or race induce disparate outcomes. For this, we sample $x_{\text{age}} \sim \mathcal{U}(18, 90)$ from a uniform distribution between 18 and 90; we sample one of three possible races $x_{\text{race}}$ and one of three possible genders $x_{\text{gender}}$ (for race and gender, the probabilities of the levels differ). Subsequently, outcome $y$ is drawn from a Bernoulli distribution conditional on $x_{\text{age}}$ and $x_{\text{race}}$. For $x_{\text{age}}$, the disparity $\rho$ is induced by a higher probability of $y=1$. Hence, $P(y=1)$ is larger for samples with a medium age as opposed to young and old persons. In addition, we induce disparities with regard to $x_{\text{race}}$. Formally, we sample 

\begin{equation}
    y \sim \text{Bernoulli}(1, {q}) \quad \text{ with } \quad {q} = \frac{f_\text{age}(x_{\text{age}}) \cdot 
    f_\text{race}(x_{\text{race}})}{g}
\end{equation}

\begin{equation}
\text{and }    f_\text{age}(x_{\text{age}}) = 
\begin{cases}
      % \vspace{5pt}
      0.5 - \rho \cdot \frac{w}{72} {,} & \text{if } x_{\text{age}} \leq 54 - \frac{w}{2} , \\
      % \vspace{5pt}
      0.5 - \rho \cdot \frac{w}{72} {,} & \text{if } x_{\text{age}} > 54 + \frac{w}{2} ,\\
      0.5 + \rho \cdot \frac{72 - w}{72} \text{,} & \text{otherwise} ,
\end{cases} 
\end{equation}

\begin{equation}
      f_\text{race}(x_{\text{race}}) = 
\begin{cases}
      0.4{,} & \text{if } x_{\text{race}} = r_1 ,\\
      0.5{,} & \text{if } x_{\text{race}} = r_2 ,\\
      0.6{,} & \text{if } x_{\text{race}} = r_3 ,\\
\end{cases} 
\end{equation}

\vspace{1pt}

where $w$ parameterizes the age interval subject to disparities. The denominator $g$ is chosen such that a valid probability distribution with $\mathbb{E}[y=1]=0.5$ is obtained. 

Synthetic dataset 1 is used to demonstrate the ability of \fairalg to (i)~automatically identify the attributes responsible for disparate outcomes (here: race and age), (ii)~handle both categorical and continuous predictors at the same time, and (iii)~accurately identify the relevant split values in continuous predictors (here: $x_\text{age} \in (54-\frac{w}{2}, 54+\frac{w}{2}]$). 

\textbf{Synthetic dataset 2:} We sample $n = 10,000$ observations with race, gender, age, and a binary outcome $y$. Here, we introduce disparities due to intersectionality. Specifically, the construction is motivated by real-world evidence of discrimination due to intersectionality \citep{Crenshaw.2015}. That is, disparate outcomes depend on the combination of both race \emph{and} gender. For this, we randomly sample two races $x_\text{race} \in \{ r_1, r_2 \}$ with equal probability, two genders $x_\text{gender} \in \{ g_1, g_2 \}$ with equal probability, and $x_{\text{age}} \sim \mathcal{U}(18, 90)$ from a uniform distribution between 18 and 90. Subsequently, the outcome $y$ is drawn from a Bernoulli distribution conditional on the combination of race \emph{and} gender via $y \sim \text{Bernoulli}(1, {s}) $
with
\begin{equation}
    {s} = 
\begin{cases}
      0.5 - \frac{\rho}{2} {,}& \text{if } (x_{\text{race}}, x_{\text{gender}}) \in \{(r_1, g_1), (r_2, g_2)\} ,\\
      0.5 + \frac{\rho}{2} {,}&
      \text{otherwise.}
\end{cases} 
\end{equation}
This implies that for samples with $(r_1, g_1)$ or $(r_2, g_2)$, the probability of a positive outcome $y = 1$ is lower than for samples with $(r_1, g_2)$ or $(r_2, g_1)$. As a result, the outcome $y$ is evenly distributed when inspecting samples by gender (or race) alone. However, disparities exist due to a combination of gender \emph{and} race.

\subsection{Real-world datasets}
\label{sec:real_dataset}

We examine disparities in two common real-world datasets from the algorithmic fairness domain (\eg, \cite{Khademi.2019, Friedler.2019, Yan.2020}):

\textbf{Adult Income:} The {Adult Income} dataset \citep{Bache.2013} is used to predict income from different sociodemographic (and potentially sensitive) attributes. Income is encoded via a binary label indicating whether that individual has an income higher than USD 50,000. Previous research suggests that there should be disparities related to gender and race (\eg, \cite{Khademi.2019, Friedler.2019, Yan.2020}). We set $\psi$ so that we measure statistical parity.

\textbf{COMPAS:} The {COMPAS recidivism} dataset \citep{Angwin.2016} includes data reoffending for criminal defendants (recidivism risk). Specifically, it includes data on the predicted vs. actual recidivism. Previous research has examined whether there are disparities across race, specifically whether there is a bias against black defendants (\eg, \cite{Angwin.2016, Friedler.2019, Black.2020, Yan.2020}). Based on the data, we construct an outcome $y_i = 0$ if the prediction for individual $i$ was correct (\ie, it matches the actual recidivism). We set an outcome $y_i = 1$ if there is a prediction error. We set $\psi$ so that we measure disparities with regard to equalized odds. Hence, we seek for individuals that are systematically sent to prison more (less) frequently despite a low (high) risk of recidivism.

\begin{figure}[t]
  \centering
  \includegraphics[width=\linewidth]{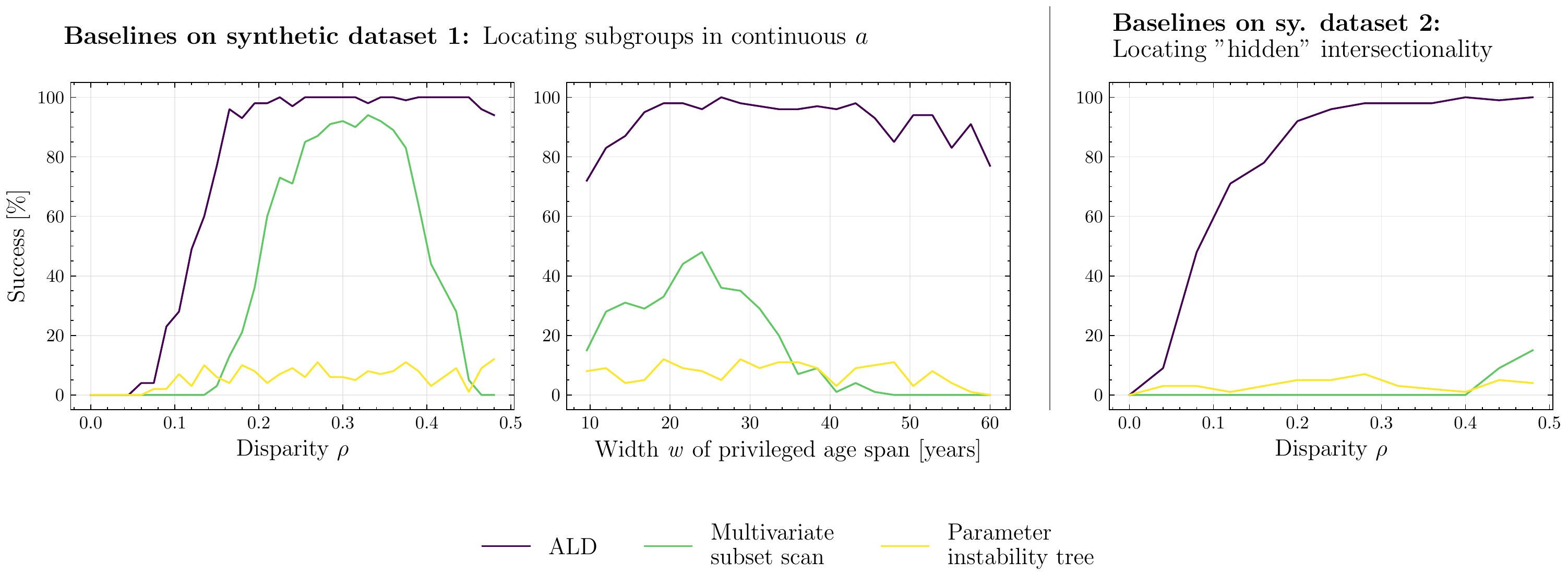}
    \caption{Performance is measured with respect to the rate (in \%) of how often the attribute causing the disparities is correctly located. Baselines are multivariate subset scan \citep{Zhang.2016} and parameter instability tree \citep{Chouldechova.2017}. Performance is averaged across 100 random draws of the synthetic datasets. In the left plot, we vary the induced disparity $\rho$ for a fixed width $w = 24 \text{ years}$ ($\frac{1}{3}$ of the total age span). In the center plot, we fix $\rho = 0.2$ and vary $w$. For the synthetic dataset 2 on the right, we vary the induced disparity $\rho$.}
  
  \Description{Plots showing the superior performance of \fairalg for locating disparate outcomes (both for continuous $a$ and in cases of intersectionality) when compared to multivariate subset scan and parameter instability tree.} % for, e.g.,  blind people
  \label{fig:res1}
\end{figure}

\section{Results}
\subsection{Results for synthetic datasets}

The results for the synthetic datasets indicate that \fairalg successfully locates complex disparities. Specifically, it succeeds both at automatically locating affected subgroups defined by continuous predictors and at revealing hidden disparities that only occur for predictor combinations, \ie, due to intersectionality.

\subsubsection{Comparison to baselines}

The results for the \textbf{synthetic dataset~1} are shown in Fig.~\ref{fig:res1} (left, center). The disparities induced by race do not prove challenging and are consistently located by all algorithms employed. Hence, in the following, the performance is measured based on the rate of how often the algorithms successfully locate disparities due to age. We count an algorithm as being successful if the located age interval does not deviate from the true age interval by more than $5\,\%$ of the total age span. 
In the left panel, the location rates are shown for varying induced disparities $\rho$. In the mid panel, the location rates are shown for different widths $w$ of the age interval.

\fairalg shows superior performance for locating disparities with regard to age. This holds for both varying $\rho$ (left) and varying $w$ (center). Multivariate subset scan (which treats age values rounded to integers as categorical) identifies the correct split values in $a$ at a considerably smaller rate and only achieves a comparable performance for $\rho \approx 0.3$. The parameter instability tree locates the correct split values in less than $15\,\%$ of runs, regardless of both $\rho$ and $w$. Overall, these results demonstrate that only \fairalg reliably locates disparities due to continuous attributes.

The results for the \textbf{synthetic dataset~2} are shown in Fig.~\ref{fig:res1} (right). Here, we measure how often the algorithms locate disparities due to a ``hidden'' combination of gender and race (intersectionality). 

\fairalg shows a superior rate at which disparity-inducing attributes are correctly located. The performance may drop for very low levels of $\rho$. Notably, both multivariate subset scan and the parameter instability tree show an inferior performance when locating disparate outcomes due to ``hidden'' intersectionality.

\subsubsection{Sensitivity analysis} We have repeated the experiments with variants of \fairalg that build upon different tree algorithms. Specifically, we have experimented with \fairalg based on traditional random forest \citep{Breiman.1984, Strobl.2009} and single trees: CART \citep{Breiman.1984}, ctree \citep{Hothorn.2006}, and CHAID \citep{Kass.1980}. However, we find that only the above variant of \fairalg (\ie, \fairalg based on a random forest of conditional inference trees \citep{Hothorn.2006, Strobl.2009}) successfully locates the complex disparities presented. The detailed results of our sensitivity analysis can be found in Appendix~\ref{app:sens_an}.

\subsection{Results for real-world datasets}

%
% Adult Income results table
%
\begin{table}[t]
    \tiny
    \centering
    \begin{tabular}{l
          >{\raggedright\arraybackslash}p{4.5cm}
          rlr}
          \hline
        rank $r$ & subgroup & \multicolumn{1}{l}{group size} & $\psi_{\text{sp}}$ & $p$  \vspace{1pt} \\
        \hline
        
        1 & \emph{race}: Asian-Pac-Islander,  Black,  White; \emph{relationship}: Husband; & $9741$ / $39.92$\,\% & $0.234$ & $<0.001$ \vspace{2pt} \\
        
        2 & \emph{marital status}: Married-AF-spouse,  Married-civ-spouse; \emph{sex}: Male; \emph{race}: Asian-Pac-Islander,  Black,  White;
        & $9842$ / $40.30$\,\% & $0.233$ & $<0.001$ \vspace{2pt} \\
        
         3 & \emph{relationship}: Husband,  Wife; \emph{sex}: Male; & $9878$ / $40.45$\,\% & $0.232$ & $<0.001$ \\
         \bottomrule
        
    \end{tabular}
    \vspace{2pt}

  \begin{tabular}{p{\columnwidth}}
  \footnotesize
  Notes: Group size refers to the absolute number of individuals within the subgroup and its relative share in the dataset. The disparity $\psi_{\text{sp}}$ refers to the statistical parity difference as defined in Eq. \ref{eq:sp} and $p$ refers to the $p$-value of the $\chi^2$-test. \fairalg is applied with $N_{\text{trees}} = 25$, $\alpha = 0.1$, and Benjamini/Hochberg correction of $p$-values.
  \end{tabular}
    
\caption{\textbf{Results for Adult Income dataset}, top 3 subgroups found for $a=\{\text{age, relationship, sex, race, marital status}\}$, ranked according to confidence. }
\label{tab:results3_a1}
\vspace{-1cm}
\end{table}

The results of \fairalg on the \textbf{Adult Income} dataset are reported in Tbl.~\ref{tab:results3_a1}. All top subgroups identified by \fairalg refer to men (\emph{sex}: Male and/or \emph{relationship}: Husband) that are married (\emph{marital status:} Married and/or \emph{relationship}: Husband). The corresponding disparities are statistically significant and imply that married men are approximately $23.5$ percentage points more likely to earn $>50,000$ USD/year than the rest of the population. On the one hand, this confirms prior research finding a strong statistical disparity with regard to gender in the Adult Income dataset (\eg, \cite{Friedler.2019, Tramer.2017}). On the other hand, \fairalg finds this disparity to be particularly pronounced for men that are married, \ie, husbands. As a consequence, a potential auditor observing the disparity with regard to gender should look more carefully at moderating factors, such as the marital status.

\begin{table*}[t]%\[!htb\]
\tiny
    \begin{subtable}[t]{.5\textwidth}
        % \caption{Ranking -- magnitude}
        \raggedright
                        \begin{tabular}[t]
            {p{0.02cm}
            >{\raggedright\arraybackslash}p{1.4cm}
                        rl 
                        >{\raggedleft\arraybackslash}p{0.56cm}
                        >{\raggedleft\arraybackslash}p{0.56cm}
                        r}
                
                \multicolumn{7}{l}{\textbf{\
                Ranking -- confidence}} \vspace{1pt} \\
                
                \hline

                $r$ & subgroup & \multicolumn{1}{l}{group size} & $\psi_{\text{eo}}$ & $\psi_{\text{fpr}}$ & $\psi_{\text{fnr}}$ & $p$  \vspace{1pt} \\
  
                \hline

                1 & AfricanAmer., NativeAmer. & $1586$ / $51.39$\,\% & $0.220$ & $0.227$ & $-0.214$ & $<0.001$  \vspace{1pt} \\
                2 & AfricanAmer. & $1581$ / $51.23$\,\% & $0.219$ & 
                $0.227$ & $-0.211$ & $<0.001$ \vspace{1pt} \\
                3 & Caucasian, Asian & $10773$ / $34.90$\,\% & $0.139$ & 
                $-0.146$ & $0.131$ & $<0.001$ \\
                \bottomrule
            \end{tabular}
    \end{subtable}%
   \begin{subtable}[t]{.5\textwidth}
   \tiny
        \raggedleft
        % \caption{Ranking -- confidence}
 \raggedright
                        \begin{tabular}[t]
            {p{0.02cm}
            >{\raggedright\arraybackslash}p{1.4cm}
                        rl 
                        >{\raggedleft\arraybackslash}p{0.56cm}
                        >{\raggedleft\arraybackslash}p{0.56cm}
                        r}
                
                \multicolumn{7}{l}{\textbf{
                Ranking -- magnitude}} \vspace{1pt} \\
                \hline
                $r$ & subgroup & \multicolumn{1}{l}{group size} & $\psi_{\text{eo}}$ & $\psi_{\text{fpr}}$ & $\psi_{\text{fnr}}$ & $p$  \vspace{1pt} \\ 
                \hline
                
                1 & Other & $168$ / $5.44$\,\% & $0.223$ & $-0.164$ & $0.282$ & $<0.001$ \vspace{1pt} \\
                2 & AfricanAmer., NativeAmer. & $1586$ / $51.39$\,\% & $0.220$ & $0.227$ & $-0.214$ & $<0.001$  \vspace{1pt} \\
                3 & AfricanAmer. & $1581$ / $51.23$\,\% & $0.219$ & 
                $0.227$ & $-0.211$ & $<0.001$ \vspace{1pt} \\
                \bottomrule
            \end{tabular}
    \end{subtable}
    \vspace{2pt}
\caption{\textbf{Results for COMPAS dataset}, top 3 subgroups found for $a=$ race, ranked according to confidence (left) and magnitude (right). Column $\psi_{\text{eo}}$ represents the magnitude and refers to the absolute odds metric (see Eq. \ref{eq:eo}). Column $\psi_{\text{fpr}}$ refers to the difference in the false positive rate, \ie, a positive $\psi_{\text{fpr}}$ implies that for individuals that will not reoffend, the probability to still be classified as \textquote{high risk} is higher if they are from this subgroup than outside of it. Analogously, $\psi_{\text{fnr}}$ refers to the difference in the false negative rate. Column $p$ represents the confidence and refers to the $p$-values from the ${\chi^2}$-tests. \fairalg is applied with $N_{\text{trees}} = 25$, $\alpha = 0.1$, and Benjamini/Hochberg correction of $p$-values.
}
\label{tab:compas_res}
\vspace{-0.5cm}
\end{table*}

The results of \fairalg on the \textbf{COMPAS} dataset are reported in Tbl.~\ref{tab:compas_res}. We apply \fairalg with $a=\text{race}$ and rank our results by both confidence (left) and magnitude (right). Again, \fairalg confirms findings of prior research \citep{Angwin.2016} by showing that ``African American'' yield a disproportionately low prediction performance. Interestingly, the ranking by magnitude (right) also hints that the comparably small subgroup with race ``Other'' may yield the lowest prediction performance, with both error rates strongly deviating from those of the rest of the population. This result gives an important hint that other racial subgroups might be affected in unexpected ways and can thus serve as a starting point for a more thorough investigation.

\section{Discussion}

\textbf{Summary of findings:} Overall, we find \fairalg to successfully locate disparities even in complex settings, namely identifying the correct split values in continuous predictors and revealing subgroups of intersectionality. For both, \fairalg outperforms existing baseline algorithms from the literature.

\textbf{Relevance:} Locating disparity-inducing attributes is highly relevant for both ethical and legal reasons. Algorithms increasingly influence the well-being of humans \citep{Martin.2019}, and are thus subject to growing attention by policy makers \citep{NYT.2016}. Indeed, recent legal initiatives aim to increase algorithmic accountability (\eg, the Algorithmic Accountability Act in the US) and fairness. For instance, the GDPR in the EU requires businesses to \textquote{ensure fair and transparent [data] processing} drawing upon \textquote{appropriate mathematical or statistical procedures} (\cite{GDPR.2016}; Article 22, Recital 71). As a consequence, algorithmic audits will become more frequent.

\textbf{Practical benefits of \fairalg:} \fairalg is designed so that key demands from ML practice in companies and organizations are met: 
\begin{enumerate}
    \item \fairalg examines arbitrary logged data. This ensures broad applicability and enables practitioners both to audit training data before fitting any model and to compare disparate outcomes of different classification models.
    \item \fairalg can be tailored to different notions of fairness. As shown above, \fairalg is effective for statistical parity and equalized odds, which are two notions that are common in practice.
    \item \fairalg can effectively locate complex disparate outcomes due to both discrete and continuous predictors. As shown in our experiments, \fairalg both accurately identifies the relevant split values in continuous predictors and successfully finds disparate outcomes due to ``hidden'' intersectionality.

\end{enumerate}

\textbf{Disclaimer on using \fairalg:} Bias, as is fairness, represents an socio-technical concept that is highly context-dependent with no universal agreed-upon definition \citep{Kleinberg.2018}. Hence, what is seen as an ``undesirable'' disparity might vary across applications, and, to arrive at such interpretations, a discourse with domain experts, as well as specialists from legal studies, ethics, and social science, is indicated. In line with this, \fairalg should only be seen as a tool for generating hypotheses of potential bias that are then inspected by domain experts. It should not be seen as a tool that provides a final disposition as to whether ML is biased. 

\textbf{Limitations:} As with any tool, \fairalg is not free of limitations that require future research. First, \fairalg can only locate disparities for sensitive attributes that are included in the data. For instance, it cannot be used to locate disparities due to sexual orientation if information on sexual orientation was not collected beforehand. This not only applies to \fairalg but all existing tools for locating and measuring disparities in the literature (\eg, \citep{Tramer.2017, Hardt.2016, Zhang.2016, Black.2020}). Here, we see the potential for future research that explores how ML audits can be conducted with missing sensitive attributes. Second, \fairalg reports outcomes from associative and not causal relationships. Hence, we caution that the output from \fairalg should not be misinterpreted as the underlying root cause due to potential hidden confounders. We emphasize that causal findings can only be established through randomized controlled trials. Here, we see further potential for future research to develop methods that can make such causal inference from observational data. Third, \fairalg does not identify where bias enters the ML pipeline. Instead, the task of \fairalg is analogous to other tools in literature where the objective is to locate and measure disparities from ML outcomes (\eg, \cite{Tramer.2017, Hardt.2016, Zhang.2016, Black.2020}). Nevertheless, \fairalg can offer substantial support to practitioners in performing ML audits.

\section{Conclusion}

There is a growing awareness that ML is not free of bias and is often responsible for disparate outcomes where one group of individuals is systematically disadvantaged. Hence, companies and organizations face the task of locating such disparities, specifically the sensitive attributes due to which a subgroup receives disparate outcomes. We propose a novel framework called \emph{Automatic Location of Disparity} (\fairalg),  which aims at identifying sources of disparities. To the best of our knowledge, \fairalg is the first framework that can locate disparities in the presence of intersectionality, that is, where complex multi-way interactions of predictors induce ML bias (and thus affect outcomes only for specific subgroups). Altogether, \fairalg offers a powerful tool to ML practitioners based on which they can perform fairness audits of their ML algorithms. This will help identifying and eventually mitigating unfairness that would otherwise disadvantage individuals, especially from marginalized groups.

\begin{acks}
Moritz von Zahn thankfully acknowledges support from efl -- the Data Science Institute.

\noindent Stefan Feuerriegel thankfully acknowledges support from the Swiss National Science Foundation (Grant 197485). 
\end{acks}

\normalsize

\bibliographystyle{ACM-Reference-Format}
\bibliography{Paper}

\clearpage
\appendix
\label{app:app}

\section{Computation of $\chi^2$-statistics}
\label{app:chi}

Once \fairalg has computed the level of disparity $\psi$, it assesses statistical significance. This is based on the $p$-values of a $\chi^2$-test, of which the characteristics depend on the notion of fairness $f$. In the case of \emph{statistical parity}, the test value $\chi^2_1$ is computed as follows:
\begin{equation}
    \chi^2_1 = 
    \frac{\abs{D}
    \big(P_{G} N_{\mathcal{X} \setminus G} - N_{G} P_{\mathcal{X} \setminus G}\big)^2}  
    {(P_{G} + N_{G})
    (N_{\mathcal{X} \setminus G} + P_{\mathcal{X} \setminus G})
    (N_{G} + N_{\mathcal{X} \setminus G})
    (P_{G} + P_{\mathcal{X} \setminus G})}
\end{equation}

where $ \abs{D} $ refers to the number of samples in the dataset, $P_G$ to the number of positive instances in subgroup $G$, $P_{\mathcal{X} \setminus G}$ to the number of positive instances outside the subgroup, $N_G$ to the number of negative instances in $G$, and $N_{\mathcal{X} \setminus G}$ defined analogously. After computing $\chi^2_1$, the $p$-value is obtained by $p\text{-value} = 1 - F(\chi^2_1)$ where $F$ is the $\chi^2_1$-distribution.

In the case of \emph{equalized odds}, two separate $\chi^2$-tests are combined, with $\chi^2_{1, \text{fpr}}$ and $\chi^2_{1, \text{fnr}}$ obtained via

\begin{equation}
\footnotesize
     \chi^2_{1, \text{fpr}} = 
    \frac{\abs{D}
    \big(FP_{G} TN_{\mathcal{X} \setminus G} - TN_{G} FP_{\mathcal{X} \setminus G}\big)^2}  
    {(FP_{G} + TN_{G})
    (TN_{\mathcal{X} \setminus G} + FP_{\mathcal{X} \setminus G})
    (TN_{G} + N_{\mathcal{X} \setminus G})
    (FP_{G} + P_{\mathcal{X} \setminus G})},
\end{equation}

\begin{equation}
\footnotesize
     \chi^2_{1, \text{fnr}} = 
    \frac{\abs{D}
    \big(FN_{G} TP_{\mathcal{X} \setminus G} - TP_{G} FN_{\mathcal{X} \setminus G}\big)^2}  
    {(FN_{G} + TP_{G})
    (TP_{\mathcal{X} \setminus G} + FN_{\mathcal{X} \setminus G})
    (TP_{G} + N_{\mathcal{X} \setminus G})
    (FN_{G} + P_{\mathcal{X} \setminus G})},
 \end{equation}
and subsequently mapped to $p_{\text{\,fpr}}$ and $p_{\text{\,fnr}}$. Next, $p_{\text{\,fpr}}$ and $p_{\text{\,fnr}}$ are combined using Fisher's method:

\begin{equation}
\footnotesize
    \chi^2_4 = -2
    \big[\log(p_{\text{\,fpr}}) + \log(p_{\text{\,fnr}})
    \big],
\end{equation}

which is then mapped to a single $p$-value.

\section{Example plots of disparate outcomes in synthetic datasets 1 and 2}
\label{app:data_plots}

\begin{figure}[h]	
	\centering
	\includegraphics[scale=1]{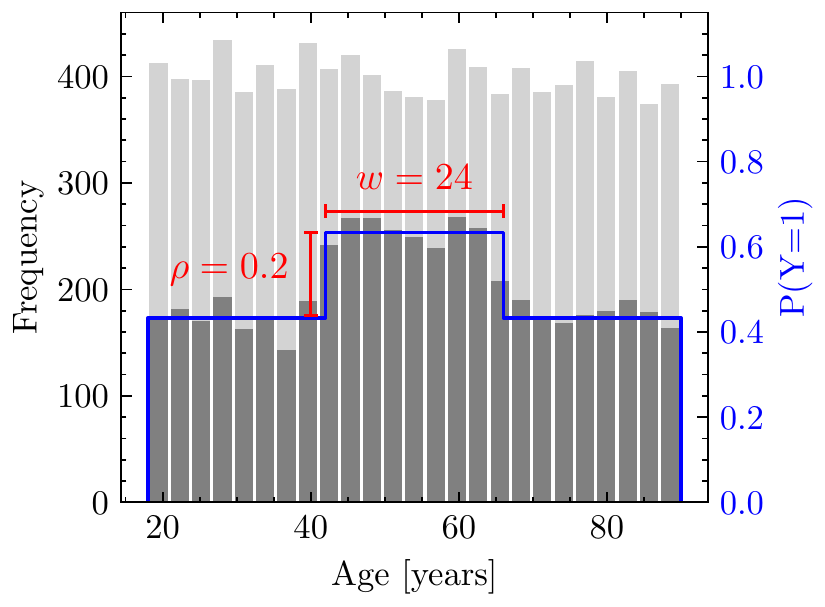}
	\caption{Probability and observed frequencies of outcome $y$ for different levels of age in a random draw of synthetic dataset~1 (light gray represents individuals with $y=0$, dark gray with $y=1$).}
	\label{fig:dataset1}
\end{figure}

\begin{figure}[h]	
	\centering
	\includegraphics[scale=0.25]{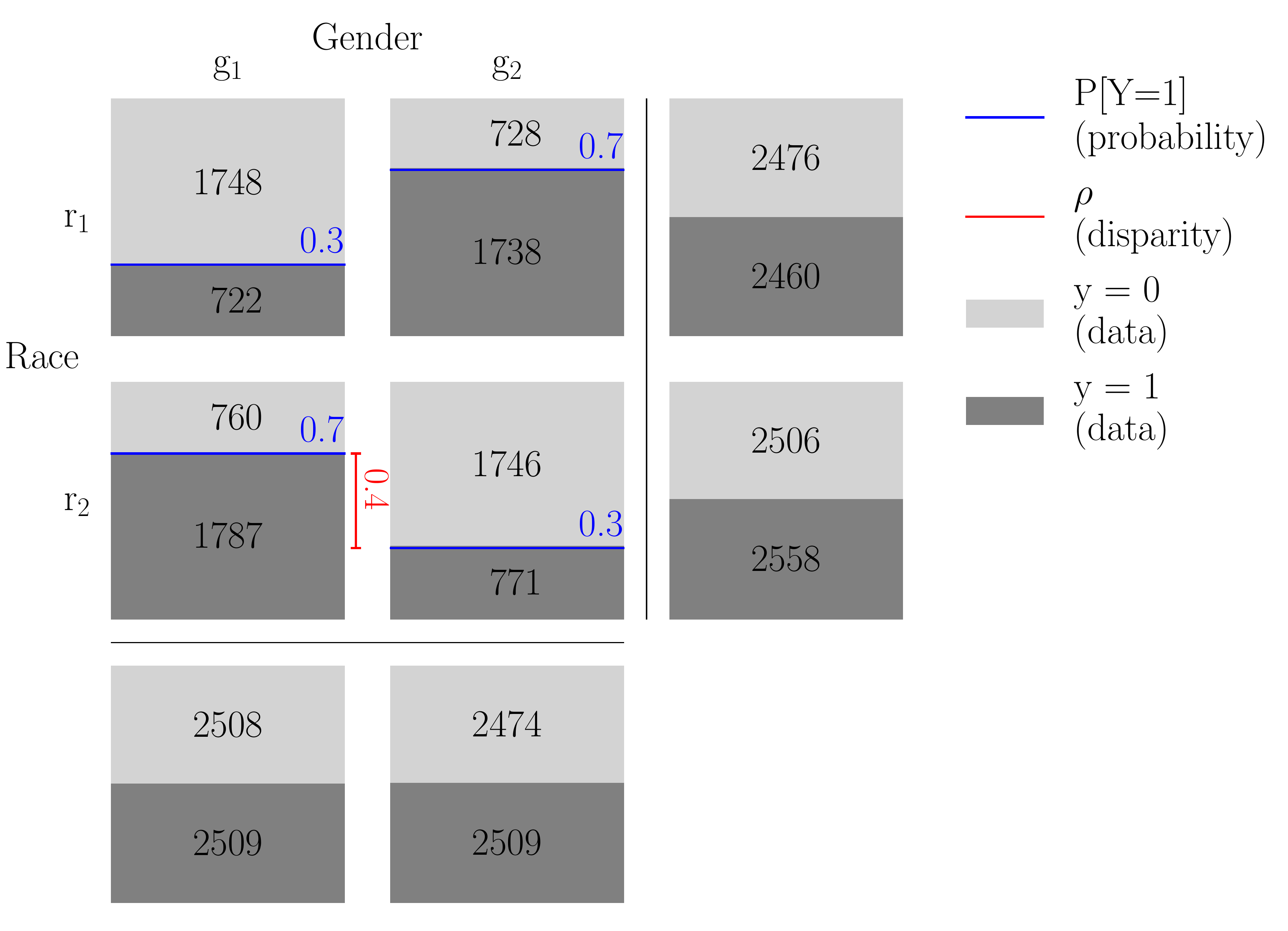}
	\caption{Probability and observed frequencies of outcome $y$ in synthetic dataset~2 with ``hidden'' intersectionality, \ie, where disparate outcomes are only induced for the combination of race and gender.}
	\label{fig:dataset2}
\end{figure}

Fig.~\ref{fig:dataset1} and Fig.~\ref{fig:dataset2} provide example plots showing  disparate outcomes in synthetic datasets 1 and 2.

\section{Sensitivity analysis}
\normalsize
\label{app:sens_an}

We implemented \fairalg via a recursive partitioning for subgroup generation. Specifically, we employed a random forest of conditional inference trees \citep{Hothorn.2006}, \ie, multiple randomized search trees perform greedy splits based on statistical confidence. Needless to say, it is possible to choose alternatives that balance exploration vs. exploitation during subgroup generation in different ways. Hence, we now report a sensitivity analysis in which we compare different variants of \fairalg. Similar to an ablation study, this allows us to assess the importance of the recursive partitioning to the overall performance.   

\begin{figure}[h]
  \centering
  \includegraphics[width=\linewidth]{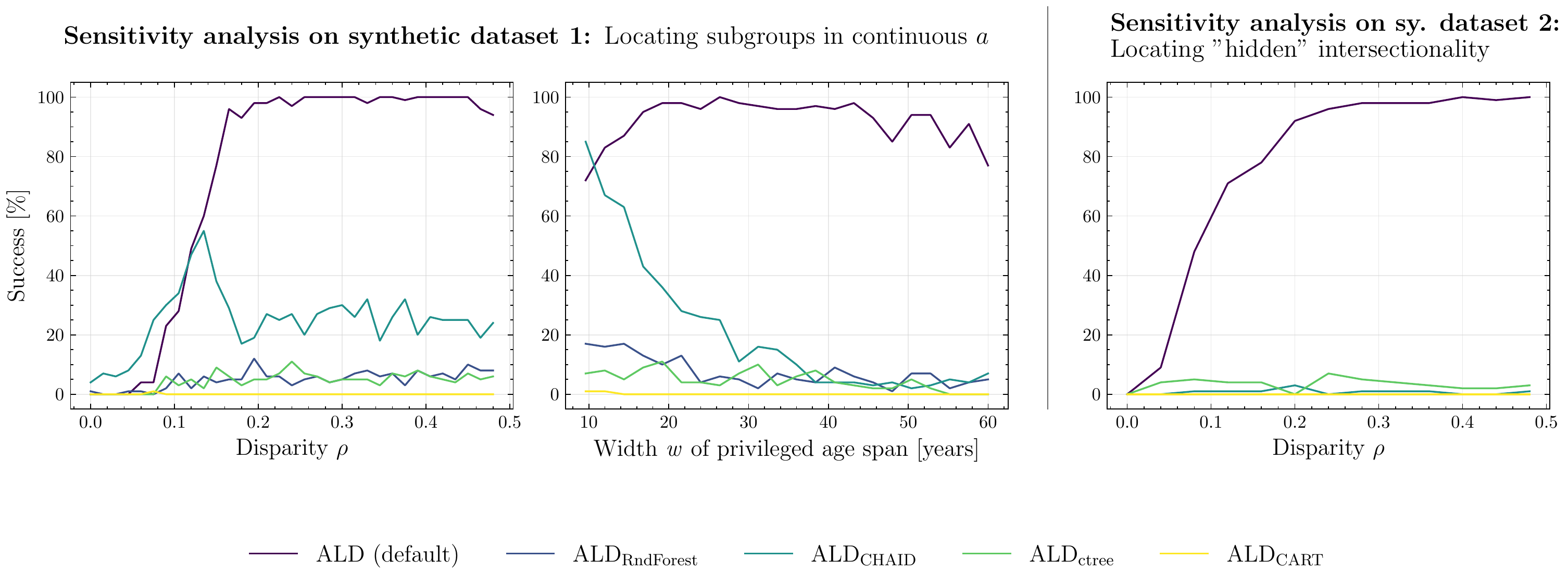}
  \caption{Performance is measured with respect to the rate (in \%) of how often the attribute causing the disparities is correctly located. Here, we compare \fairalg with different tree-based algorithms. Results are based on 100 random draws of synthetic datasets~1 (left, center) and~2 (right). On the left, we vary the induced disparity $\rho$ for a fixed width $w = 24 \text{ years}$ ($\frac{1}{3}$ of the total age span). In the center, we fix $\rho = 0.2$ and vary $w$. For the synthetic dataset 2 on the right, we vary the induced disparity $\rho$.}
  \Description{Plots showing the superior performance of \fairalg with conditional inference forest for locating disparate outcomes (both for continuous $a$ and in cases of intersectionality) when compared to \fairalg with other search trees.} % for, e.g.,  blind people
  \label{fig:res2}
%  \vspace{-0.5cm}
\end{figure}

\textbf{Variants of \fairalg:} We implement several variants of our standard \fairalg. Therein, we replace the recursive partitioning with algorithms for recursive partitioning:
\begin{enumerate}[leftmargin=0.5cm]
\item \fairalg (default): The ``original'' \fairalg uses conditional inference forest \citep{Hothorn.2006, Strobl.2009}. This allows us to to construct multiple randomized trees. We made good experience with changing two default values and setting $N_\text{trees} = 25$ and min. criterion $\alpha=0.1$.
\item \fairalg$_\text{ctree}$: We implement a \fairalg variant based on a single ctree \citep{Hothorn.2006}. We compare  \fairalg to \fairalg$_\text{ctree}$ in order demonstrate the importance of having multiple randomized ctrees.
\item \fairalg$_\text{CHAID}$: We implement Chi-square Automatic Interaction Detector (CHAID \cite{Kass.1980}) as another single search tree based on statistical testing. CHAID cannot handle continuous predictors, so that we discretize ``age'' by integers and consider it as an ordinal predictor.
\item \fairalg$_\text{CART}$: We implement an \fairalg variant based on CART \citep{Breiman.1984}. CART was chosen due to its widespread use in subgroup analyses but in other contexts (\eg, \cite{Su.2009}). CART \citep{Breiman.1984} performs a recursive partitioning through a tree algorithm.  Specifically, CART splits variables so that the purity of the outcome within the nodes is maximized. This greedy approach is similar to the association-guided tree construction by \citet{Tramer.2017} (however, the latter tree construction does not allow one to perform splits for predictors, and, hence, direct adaptations for our purpose are not possible). In our implementation of \fairalg$_\text{CART}$, we additionally employ well-known procedures to prevent overfitting. Specifically, we bound the maximum depth of trees to $8$ and prune nodes representing fewer than $100$ instances (\ie, 1\,\% of the input data). 
\item \fairalg$_\text{RndForest}$: For comparison, we additionally implement the traditional random forest based on CART \citep{Breiman.1984, Strobl.2009}. Here, we set $N_\text{trees}=25$, max\_depth$=8$, and min\_samples\_leaf$=100$.
\end{enumerate}

\noindent
\textbf{Results:} The results are shown in Fig.~\ref{fig:res2}. The results demonstrate the superiority of \fairalg based on conditional inference forest. For synthetic dataset 1, \fairalg (\ie, based on conditional inference forest) shows superior performance over variants with traditional random forest or single trees. Only for subgroups exhibiting a low level of induced disparity $\rho$ (Fig.~\ref{fig:res2}; left) or defined over a very small interval width $w$ (Fig.~\ref{fig:res2}; center), \fairalg$_\text{CHAID}$ outperforms \fairalg. However, looking at the performance over the whole range of $w$ and $\rho$, \fairalg performs considerably better. For synthetic dataset 2, the picture is even clearer: only \fairalg successfully locates disparate outcomes even in the presence of ``hidden'' intersectionality (Fig.~\ref{fig:res2}; right).

\end{document}